\documentclass[runningheads]{llncs}
\usepackage[T1]{fontenc}
\usepackage{graphicx}
\usepackage{subcaption}
\usepackage{multirow}
\usepackage{makecell}
\usepackage{booktabs}
\usepackage{amsmath} 
\usepackage{hyperref}
\usepackage{tikz}
\usetikzlibrary{positioning, shapes, arrows.meta}

\begin{document}
\title{A Comparative Study of Decoding Strategies in Medical Text Generation}

\author{Oriana Presacan\inst{1}\orcidID{0009-0005-6909-4476} \and
Alireza Nik\inst{2,3}\orcidID{0009-0001-3551-3215} \and
Vajira Thambawita\inst{2}\orcidID{0000-0001-6026-0929} \and
Bogdan Ionescu\inst{1}\orcidID{0009-0005-0535-1306} \and
Michael Riegler\inst{4}\orcidID{0000-0002-3153-2064}}

\authorrunning{O. Presacan et al.}

\institute{AI Multimedia Lab, CAMPUS Research Institute, National University of Science and Technology Politehnica Bucharest, 060042 Bucharest, Romania  \and
Department of Holistic Systems, SimulaMet, 0170 Oslo, Norway\\ \and
Oslo Metropolitan University, 0176 Oslo, Norway\\ \and
Cyber Security, Simula Research Laboratory, 0164 Oslo, Norway\\
}

\maketitle         

\begin{abstract}
Large Language Models (LLMs) rely on various decoding strategies to generate text, and these choices can significantly affect output quality. In healthcare, where accuracy is critical, the impact of decoding strategies remains underexplored. We investigate this effect in five open-ended medical tasks, including translation, summarization, question answering, dialogue, and image captioning, evaluating 11 decoding strategies with medically specialized and general-purpose LLMs of different sizes. Our results show that deterministic strategies generally outperform stochastic ones: beam search achieves the highest scores, while $\eta$ and top-$k$ sampling perform worst. Slower decoding methods tend to yield better quality. Larger models achieve higher scores overall but have longer inference times and are no more robust to decoding. Surprisingly, while medical LLMs outperform general ones in two of the five tasks, statistical analysis shows no overall performance advantage and reveals greater sensitivity to decoding choice. We further compare multiple evaluation metrics and find that correlations vary by task, with MAUVE showing weak agreement with BERTScore and ROUGE, as well as greater sensitivity to the decoding strategy. These results highlight the need for careful selection of decoding methods in medical applications, as their influence can sometimes exceed that of model choice.


\vspace{-5pt}
\end{abstract}

\section{Introduction}
\vspace{-5pt}

In recent years, we have witnessed a growing interest in adopting generative Artificial Intelligence (AI), including Large Language Models (LLMs), in healthcare applications \cite{he2025survey,nori2023,tian202}. Medical tasks ranging from decision making and question answering to translation can now be carried out using state-of-the-art models, such as OpenAI's GPT-4 \cite{gpt4} and Google's Gemini \cite{gemini}. These models have shown impressive potential in the analysis, interpretation, and generation of high-quality responses in medical exams \cite{nori2023,stribling2024}. Considering the sensitive nature of the medical domain, ensuring that these AI models generate medically accurate responses in real-world situations is a key challenge.  

An important component affecting the quality, diversity, and fluency of the generated LLM responses is the decoding strategy used during inference. LLMs generate text in an autoregressive manner, sequentially predicting the next token conditioned on all previously generated tokens in the sequence \cite{vaswani2023attentionneed}. At each step, the model computes a probability distribution over the vocabulary and selects the next token accordingly. The simplest approach, greedy decoding \cite{wiher2022decoding}, always chooses the token with the highest probability. While this deterministic method can produce coherent and consistent text, it often leads to “flat” and repetitive outputs, limiting diversity. In contrast, stochastic decoding methods, such as top-k \cite{fan2018topk}, top-p \cite{holtzman2020topp}, etc., introduce controlled randomness, which can yield more dynamic, engaging, and varied responses. Decoding strategies, therefore, involve a trade-off between the familiar and the unexpected, balancing predictability with creativity \cite{zhang2020trading}. In domains such as healthcare, this balance is particularly important: some deterministic methods may ensure precision but risk generating generic or overly rigid content, whereas overly stochastic methods may enhance contextual richness but at the cost of factual reliability. Selecting an appropriate decoding strategy is thus a trade-off between generating medically precise, consistent information and producing diverse, contextually rich text, which are essential for tasks like medical dialogue and summarization \cite{calle2024towards,xie2023faithful}.

The current literature shows that decoding techniques in general text generation are rapidly evolving \cite{wangbeyond}. Research has progressed from deterministic \cite{freitag2017beam,vijayakumar2018diverse} and stochastic methods \cite{fan2018topk,hewitt2022eta,holtzman2020topp} to more advanced strategies aimed at improving the factuality and coherence of the output text \cite{chuang2024dola,Su2022contrastive}, or even reducing decoding latency \cite{leviathan}. Although the effect of decoding methods has recently been studied on general open-ended benchmarks \cite{arias2024decoding,shi2024thorough}, their impact remains underexplored in the medical domain. As new medical benchmarks are introduced \cite{yan2024}, most research has focused on model architecture, overlooking decoding strategies despite their influence on generative quality. Furthermore, the role of decoding methods in medical-domain LLMs, such as models fine-tuned or pre-trained for clinical or biomedical tasks, has received even less attention, leaving a gap in understanding how these techniques affect performance in specialized models.

The main contribution of this research is to investigate the impact of various decoding strategies on LLM performance in the medical domain, including: 
\vspace{-5pt}

\begin{itemize}
    \item \small\textbf{Comprehensive analysis of various decoding strategies in healthcare:} we systematically study how different decoding strategies influence the generative performance of general-purpose and medical LLMs on medical datasets.
    \item \small\textbf{Task-specific evaluation of decoding methods:} we investigate which decoding strategies are most effective and efficient for different types of medical tasks.
    \item \small\textbf{Model performance and sensitivity to decoding strategies:} we analyze the performance and sensitivity of various LLMs, ranging from small to large, general-purpose to medical models, to changes in decoding configurations. 
    \item \small\textbf{Evaluation metrics sensitivity and agreement:} we assess the stability and sensitivity of the evaluation metrics (BLEU, ROUGE, BERTScore, and MAUVE) in different tasks, and their agreement with each other.
    \item \small\textbf{Computational efficiency analysis:} We report the inference time across decoding strategies and models, offering practical insights into their runtime trade-offs, which are beneficial for resource-constrained healthcare applications.
\end{itemize}

The remainder of the paper is organized as follows: Section 2 provides background information on the decoding strategies evaluated in this study. Section 3 describes the implementation details, including the logic behind LLM and metric selection, hyperparameters, benchmarks, and hardware. Section 4 presents the results and statistical analyses, followed by a discussion of the findings in Section 5. Finally, Section 6 concludes the paper. The code for text generation and statistical analysis, along with the complete set of result tables, is available at our \href{https://github.com/orianapresacan/decoding_healthcare}{GitHub repository}.

\section{Background}
\subsection{Decoding Strategies} 


\subsubsection{Deterministic Strategies}
\paragraph{\textbf{Greedy}} \cite{wiher2022decoding} method is the simplest strategy as the model always chooses the token with the highest probability. However, it limits the diversity and can lead to repetitive phrasing.
\vspace{-5pt}

\paragraph{\textbf{Beam Search (BS)} \cite{freitag2017beam}} method keeps the top-k most probable sequences, where k is a user-defined hyperparameter. At each decoding step, it expands the sequence by all possible next tokens, calculates their cumulative probabilities, and retains the top-k sequences. Unlike greedy decoding, BS aims to find a globally optimal sequence rather than making the best choice at each step.
\vspace{-5pt}

\paragraph{\textbf{Diverse Beam Search (DBS)} \cite{vijayakumar2018diverse}} decoding strategy enhances BS by introducing more diversity among candidate outputs. It does this by dividing the beams into a user-defined number of groups and adding a diversity penalty to reduce overlap between groups.
\vspace{-5pt}

\paragraph{\textbf{Contrastive Search (CS)} \cite{Su2022contrastive}} is an extension of BS that aims to improve text quality by reducing repetition. It selects the next token based on both model confidence and a semantic similarity penalty. It uses two hyperparameters: \(k\) (top-\(k\) candidates) and \(\alpha\) (penalty weight), to balance fluency and diversity.
\vspace{-5pt}

\paragraph{\textbf{DoLa}~\cite{chuang2024dola}} takes a different approach by using outputs from both the final layer and an earlier layer of the model. It compares these outputs, giving more weight to the final one to guide generation. This helps highlight tokens that are consistently supported across layers.

\subsubsection{Stochastic Strategies}
\paragraph{\textbf{Temperature Sampling} \cite{ackley1985learning}} strategy rescales the logits before applying the softmax function. A lower temperature sharpens the distribution, favoring high-probability tokens, while a higher temperature flattens it, increasing the likelihood of selecting lower-probability tokens and thereby enhancing output diversity. Temperature sampling can be combined with any of the following stochastic strategies; in our experiments, we fix $T=1$ when evaluating the other strategies.

\paragraph{\textbf{Top-k Sampling} \cite{fan2018topk}} limits the sampling process to the top-k most probable tokens. The next token is then sampled only from this restricted subset, where k is a predefined hyperparameter.
\vspace{-5pt}

\paragraph{\textbf{Top-p (nucleus) Sampling} \cite{holtzman2020topp}} strategy selects the smallest set of top-ranked tokens whose cumulative probability reaches a threshold p, allowing the number of candidates to vary with the model’s confidence rather than being fixed as in top-k sampling.
\vspace{-5pt}

\paragraph{\textbf{\(\eta\)-Sampling} \cite{hewitt2022eta}} strategy keeps only tokens above a probability threshold that adapts based on the entropy of the token distribution. When the model is confident (low entropy), only top tokens are kept. When it's uncertain (high entropy), more tokens are included for flexibility.
\vspace{-5pt}

\paragraph{\textbf{Min-p Sampling} \cite{nguyen2025minpsampling}} sampling relies solely on the most probable token, while \(\eta\)-sampling adapts its threshold based on the overall entropy of the distribution. It sets the cutoff relative to this top token’s probability, scaling it by a hyperparameter to determine the sampling pool. By adapting directly to the model’s confidence in its best prediction, min-p remains stable even at high temperatures, where other sampling methods often produce more chaotic outputs.
\vspace{-5pt}

\paragraph{\textbf{Typical Sampling} \cite{meister2023typical}} method selects tokens closest to the average entropy of the distribution. A threshold hyperparameter controls how much deviation from this average is allowed, filtering out tokens that are either too predictable or too unlikely, and focusing on those that are most typical for the given context.

\section{Implementation Details}
\setcounter{footnote}{0}

\subsection{LLM selection}

We adopted a systematic approach to the selection of LLMs, focusing exclusively on open-source models and imposing a maximum size limit of 14 billion parameters due to computational constraints. The selection covered three categories: general-purpose, medical, and multimodal models. For general-purpose LLMs, we grouped candidates by model size (small, medium, and large) and selected representative models within each group. All models were evaluated consistently across tasks using greedy decoding, and for each task, we retained the best-performing model within each category and size group. Medical-specific models were selected based on their prominence in recent biomedical NLP literature and availability in open-source form. For multimodal tasks, we evaluated several vision-language models, including both general-purpose and medical-specialized variants. Figure \ref{fig:model-selection} presents an overview of the selected models, grouped by type and size, as well as the final model choices per task.
\begin{figure}[ht!]

\noindent{\textbf{(A)}}

\vspace{0.5em}

 \centering
\begin{tikzpicture}[
  node distance=0.2cm and 0.6cm,
  box/.style={rectangle, draw, rounded corners, fill=gray!5, text width=5cm, align=left, inner sep=8pt},
]

\node[box] (general) {
  \textbf{General-Purpose Models} \\[3pt]
  \textbf{Small (1–2B):}
  \begin{itemize}
    \item Qwen3-1.7B
    \item gemma-3-1b-it
    \item Llama-3.2-1B-Instruct
  \end{itemize}
  \textbf{Medium (7–8B):}
  \begin{itemize}
    \item Qwen3-8B
    \item Mistral-7B-Instruct-v0.3
    \item Llama-3.1-8B-Instruct
  \end{itemize}
  \textbf{Large (12–14B):}
  \begin{itemize}
    \item Qwen3-14B
    \item gemma-3-12b-it
    \item Llama-2-13b-chat-hf
  \end{itemize}
};

\node[box, right=of general.north, anchor=north, xshift=5.2cm] (medical) {
  \textbf{Medical-Specific Models} \\[3pt]
  \begin{itemize}
    \item BioMistral-7B
    \item meditron-7B
    \item medalpaca-7B
    \item medgemma-4B-it
  \end{itemize}
};

\node[box, below=of medical] (multimodal) {
  \textbf{Multimodal Models} \\[3pt]
  \textbf{Medical:}
  \begin{itemize}
    \item medgemma-4b-it
    \item llava-med-v1.5-mistral-7b
  \end{itemize}
  \textbf{General:}
  \begin{itemize}
    \item gemma-3-12b-it
    \item llava-1.5-7b-hf
  \end{itemize}
};
\end{tikzpicture}

\noindent\hspace*{-35em}{\textbf{(B)}}

\renewcommand{\arraystretch}{1.3}
\vspace{1em}
\small  
\begin{tabular}{l@{\hskip 8pt}l@{\hskip 8pt}c@{\hskip 8pt}c@{\hskip 8pt}c@{\hskip 8pt}c@{\hskip 8pt}c}
\toprule
\textbf{Type} & \textbf{Size} & \textbf{Trans} & \textbf{Summ} & \textbf{QA} & \textbf{Dialogue} & \textbf{Images} \\
\midrule
\multirow{3}{*}{\textbf{General}} 
  & Small  & Qwen    & Qwen     & Llama      & Qwen      & -- \\
  & Medium & Qwen    & Llama    & Mistral    & Qwen      & -- \\
  & Large  & Qwen    & Qwen     & Llama      & Qwen      & -- \\
\midrule
\multirow{1}{*}{\textbf{Medical}} 
  & --     & medalpaca & medgemma & BioMistral & medgemma & -- \\
\midrule
\multirow{2}{*}{\textbf{Multimodal}} 
  & General & -- & -- & -- & -- & llava \\
  & Medical & -- & -- & -- & -- & llava-med \\
\bottomrule
\end{tabular}

\caption{Panel A: All evaluated LLMs, categorized by type and size, with their Hugging Face model names. Panel B: Final selected models per task.}
\label{fig:model-selection}
\end{figure}

\subsection{Hyperparameters}
We explored a range of hyperparameters for each method and selected those that yielded the best performance for our final analysis. The hyperparameter values were informed by prior literature \cite{shi2024thorough} and are summarized in Table \ref{tab:Table1}. 

\begin{table}[h!]
\centering
\begin{tabular}{|l|l|l|}
\hline
\textbf{Strategies} & \textbf{Hyperparameter} &\textbf{Values} \\
\hline
BS & beam size &\{3, 5, 10\} \\ \hline 
DBS & (beam size, diversity penalty)  &\{(4,2), (4,4), (8,2), (8,4)\} \\ \hline 
CS & penalty $\alpha$, top-k candidate pool size &$\{\alpha$: \{0.1, 0.2, 0.3, 0.4, 0.5, 0.6\}, $k$= 6\} \\ \hline 
Dola & dola layer &\{low, high\} \\ \hline 
temperature & temperature &\{0.3, 0.5, 0.7, 0.9\} \\ \hline 
top p & nucleus sampling probability threshold &\{0.8, 0.85, 0.9, 0.95\} \\ \hline 
top k & candidate pool size &\{5, 25, 50, 100\} \\ \hline 
min p & minimum token probability threshold &\{0.05, 0.1, 0.3, 0.5\} \\ \hline 
eta & $\eta$-sampling cutoff threshold &\{0.0003, 0.0006, 0.0009, 0.002, 0.004\} \\ \hline 
typical & typicality threshold &\{0.2, 0.5, 0.9, 0.95\} \\ 
\hline
\end{tabular}
\caption{Ranges of hyperparameter values tested for the decoding strategies.}
\label{tab:Table1}
\vspace{-5pt}
\end{table}

\subsection{Medical Benchmarks} 
We selected five different open-ended medical tasks, widely used in healthcare: translation, summarization, question answering, dialogue, and image captioning. For each, we used only 100 samples, due to the experimental nature of this study and constraints on available resources.

For the medical translation task, we used the UFAL Medical Corpus 1.0\footnote{\url{https://ufal.mff.cuni.cz/ufal_medical_corpus}}, which contains medical text translations in multiple languages. We focused on the German–English pairs and applied preprocessing to retain only medical-domain entries, remove duplicates, and filter out short or low-quality sentences based on length, numeric content, and symbolic artifacts. The final dataset was uniformly sampled across sources.

Pubmed-summarization dataset \cite{summarization}, which consists of full-text biomedical articles from PubMed paired with their corresponding abstracts, was used for the summarization task. 

The medical QA dataset\footnote{\url{https://huggingface.co/datasets/medalpaca/medical_meadow_medical_flashcards}} we selected contains open-ended medical question–answer pairs covering topics like anatomy, physiology, and pharmacology. It was derived from flashcards created by students and the content was cleaned for use in training and evaluating medical QA models.

For open-ended healthcare dialogues, we used Healthbench\footnote{\url{https://openai.com/index/healthbench/}}, a benchmark by OpenAI to assess LLMs in a range of real-world medical tasks. The benchmark includes multi-turn dialogues between LLMs and individuals, evaluated using a rubric designed by medical professionals.
However, in our study, we only considered the physicians' completions of the dialogue as the ground truth and did not apply the rubric-based evaluation. We randomly selected single-turn dialogues with the global health theme.

Lastly, for image captioning, we used ROCOv2 (Radiology Objects in Context Version 2) \cite{rocov2}, a multimodal dataset consisting of medical images and corresponding figure captions. It was derived from the PubMed Open Access subset and includes radiology images across various anatomical regions and seven different imaging modalities.

\subsection{Metric Selection}
To enable consistent comparisons across tasks, we applied two common metrics to all benchmarks: ROUGE, a traditional metric, and BERTScore, a newer embedding-based metric. Additionally, we used BLEU for translation and MAUVE for dialogue and QA. This approach allowed us to evaluate individual task performance while also identifying trends across models and metrics.

\subsection{Hardware and Software Setup}
We conducted all experiments on a machine equipped with NVIDIA A100-PCIE-40GB GPUs and a dual 64-core AMD EPYC 7763 CPU @ 2.45GHz. Our implementation used Python 3.10.12 and PyTorch 2.6.0+cu126. Model loading and inference were handled using the Hugging Face \textit{Transformers}\footnote{\url{https://huggingface.co/docs/transformers/en/index}} library, and evaluation metrics were computed using the \textit{Evaluate}\footnote{\url{https://huggingface.co/docs/evaluate/en/index}} library from Hugging Face.
\vspace{-10pt}

\section{Evaluation Results}
\vspace{-5pt}
In our analysis, we evaluated 11 decoding strategies across 5 medical tasks, using 4 LLMs for each benchmark (and 2 for the multimodal task). Each configuration was assessed with 2-3 evaluation metrics, in addition to inference time, resulting in a total of 718 observations (with CS results missing for MedGemma due to a technical incompatibility encountered during implementation). Given the non-normal distribution of our data, we applied non-parametric statistical methods. The following subsections present our findings, organized into three categories: decoding strategies, models, and evaluation metrics. 

\subsection{Decoding Strategies}
Table~\ref{tab:Table2} reports the results for the top-performing LLMs across benchmarks; the complete set of results is provided in our \href{https://github.com/orianapresacan/decoding_healthcare}{repository}. The impact of decoding strategies differs by task: largest in dialogue and QA and moderate in translation, summarization, and captioning. Overall, BS had the best mean rank, followed by CS and DBS; top-p, eta, and top-k performed the worst.

\begin{table}[h]
\centering
\begin{tabular}{|l|c|c|c|c|c|}
\hline
\multirow{3}{*}{} 
& \textbf{Translation} & \textbf{Summarization} & \textbf{QA} & \textbf{Dialogue} & \textbf{Captioning} \\
& Qwen 14B & Llama 8B & MedAlpaca 7B & Qwen 14B & Llava Med \\
 & BLEU & ROUGE & ROUGE & MAUVE & ROUGE \\
\hline
greedy & 0.4422 & 0.2252 & 0.3839 & 0.9141 & 0.1725 \\ \hline
BS & 0.4379 & \textbf{0.2374} & \textbf{0.4197} & 0.8870 & 0.1718 \\ \hline
DBS & 0.4388 & 0.2329 & 0.3578 & 0.8196 & 0.1715 \\ \hline
CS & \textbf{0.4428} & 0.2305 & 0.3845 & \textbf{0.9379} & 0.1723 \\ \hline
DoLa & 0.4285 & 0.2263 & 0.3798 & 0.8967 & 0.1632 \\ \hline
temperature & 0.4379 & 0.2255 & 0.3617 & 0.9322 & 0.1645 \\ \hline
top p & 0.4415 & 0.2217 & 0.3354 & 0.9309 & 0.1590 \\ \hline
top k & 0.4392 & 0.2162 & 0.3137 & 0.8776 & 0.1543 \\ \hline
min p & 0.4419 & 0.2272 & 0.3649 & 0.9252 & \textbf{0.1756} \\ \hline
eta & 0.4392 & 0.2181 & 0.2972 & 0.9158 & 0.1547 \\ \hline
typical & 0.4415 & 0.2223 & 0.3377 & 0.9080 & 0.1577 \\
\hline
\end{tabular}
\caption{Results across decoding strategies and tasks for the best-performing LLM; bold values indicate the highest result for that model.}
\label{tab:Table2}
\vspace{-5pt}
\end{table}

To assess whether there are statistically significant differences between the decoding strategies, we used ROUGE as a consistent evaluation metric across tasks. Scores were normalized within each task to account for scale differences. We then applied the \textit{Friedman} test, a non-parametric statistical test suitable for comparing multiple treatments (decoding strategies) across repeated measures (task$\times$model blocks). The test was significant ($\chi^2=44.25$, $df=10$, $p=2.97\times10^{-6}$; $n=16$ blocks), indicating that at least one strategy performs differently from the others and that further tests are needed to identify where the differences lie. Therefore, we ran pairwise \textit{Wilcoxon} signed-rank tests across task$\times$model blocks. Unlike the \textit{Friedman} test, which only detects overall differences, \textit{Wilcoxon} enables direct comparisons between strategy pairs. We applied the \textit{Holm} correction to control the family-wise error rate, which is important when making multiple pairwise comparisons. This analysis revealed significant differences between:

\begin{itemize}
    \item BS $>$ eta ($W= 7.5$, $p_{\mathrm{Holm}}=0.011$, $\Delta$rank = 6.50, $n=18$);
    \item BS $>$ top-$p$ ($W=10$, $p_{\mathrm{Holm}}=0.017$, $\Delta$rank = 4.75, $n=18$);
    \item BS $>$ top-$k$ ($W=11$, $p_{\mathrm{Holm}}=0.022$, $\Delta$rank = 7.00, $n=18$);
    \item DBS $>$ eta ($W=7.5$, $p_{\mathrm{Holm}}=0.011$, $\Delta$rank = 4.75, $n=18$);
    \item DBS $>$ top-$k$ ($W=11$, $p_{\mathrm{Holm}}=0.022$, $\Delta$rank = 5.25, $n=18$);
\end{itemize}
where $W$ is the \textit{Wilcoxon} test statistic, $p_{\mathrm{Holm}}$ is the \textit{Holm}-adjusted p-value, $\Delta$ is the median rank difference between strategies, $n$ is the number of paired samples (task$\times$model blocks).

While the above analysis was conducted across all tasks, we now examine each task individually to determine whether certain decoding strategies are better suited for specific tasks. Based on average ranks across models, DBS performed best for translation, min p sampling for summarization, CS for image captioning, and BS for dialogue and QA. When applying the \textit{Friedman} test per task, we found statistically significant differences between strategies only in QA and dialogue. However, paired \textit{Wilcoxon} signed-rank tests found no statistically significant differences between individual strategy pairs, suggesting that, although some strategies appear to perform better on average, the differences are not strong enough within individual tasks.

Next, we grouped strategies as deterministic vs.\ stochastic and compared the groups. A paired \textit{Wilcoxon} signed-rank test on mean differences showed a small but statistically significant advantage for deterministic decoding (median paired difference $\Delta\text{ROUGE}=0.0029$, $p=0.0003$, $n=18$, $W=10$, $n=18$). At the per-task level ($n=4$ blocks each), tests were not significant, but the medians consistently favored deterministic decoding.

Lastly, we also analyzed how decoding strategies impact inference time (measured in seconds per token) and found significant differences across methods (\textit{Friedman test}: $\chi^2=125.33$, $p<0.001$). Top-p, top-k, typical were the fastest, while BS, CS, DBS were the slowest. Additionally, we found a statistically significant positive correlation between the normalized inference time and ROUGE performance (Kendall’s $\tau = 0.4909$, $p = 0.0405$), suggesting that, on average, slower decoding strategies tend to yield better output quality.


\subsection{Models}
\begin{figure*}[ht!]
  \centering
  \begin{subfigure}[t]{0.48\textwidth}
    \centering
    \includegraphics[width=\linewidth]{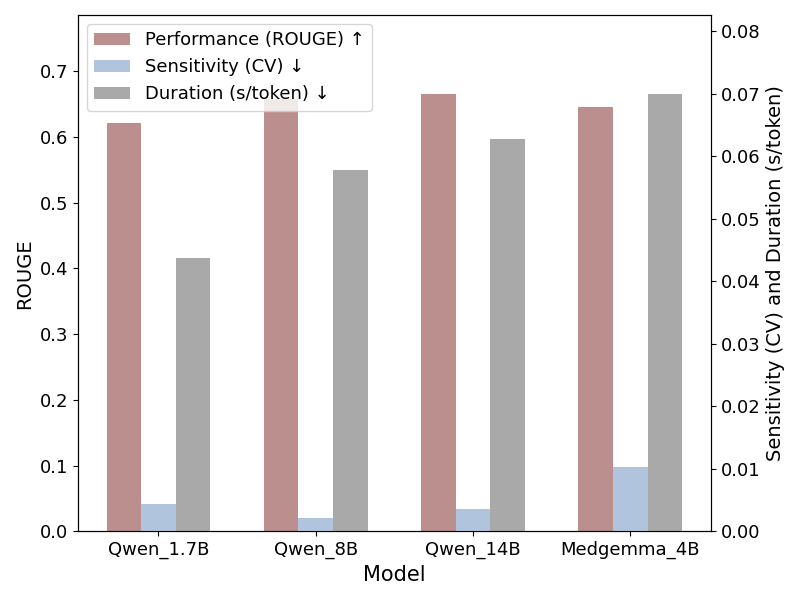}
    \caption{Translation}
    \label{fig:task1}
  \end{subfigure}\hfill
  \begin{subfigure}[t]{0.48\textwidth}
    \centering
    \includegraphics[width=\linewidth]{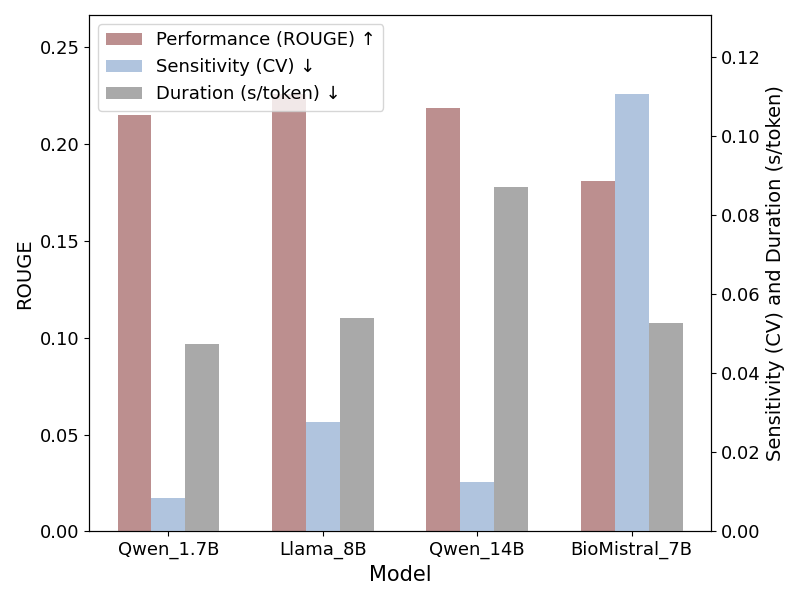}
    \caption{Summarization}
    \label{fig:task2}
  \end{subfigure}\hfill
  \begin{subfigure}[t]{0.48\textwidth}
    \centering
    \includegraphics[width=\linewidth]{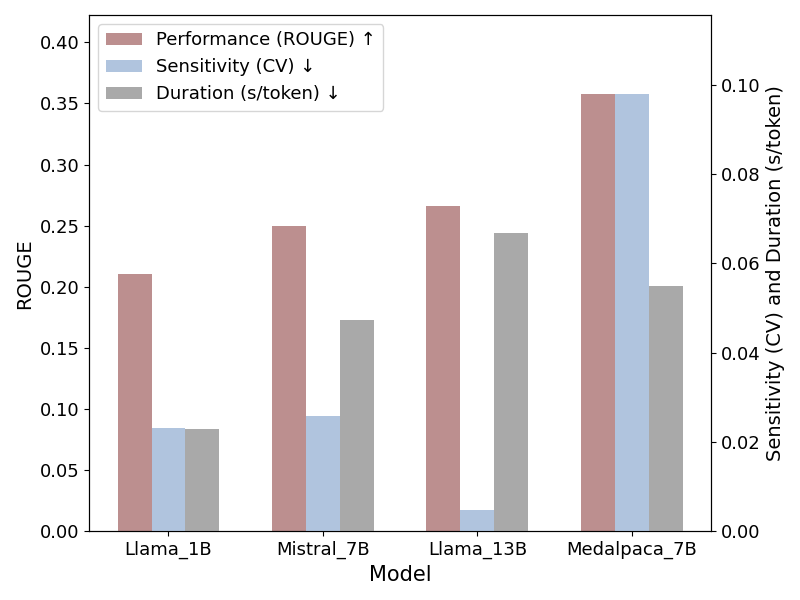}
    \caption{QA}
    \label{fig:task3}
  \end{subfigure}
  \begin{subfigure}[t]{0.48\textwidth}
    \centering
    \includegraphics[width=\linewidth]{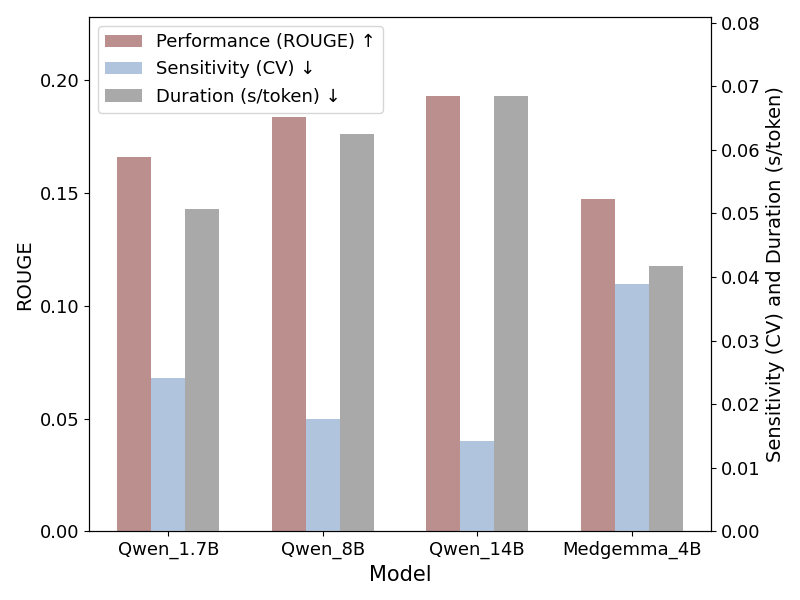}
    \caption{Dialogue}
    \label{fig:task4}
  \end{subfigure}\hfill
  \begin{subfigure}[t]{0.48\textwidth}
    \centering
    \includegraphics[width=\linewidth]{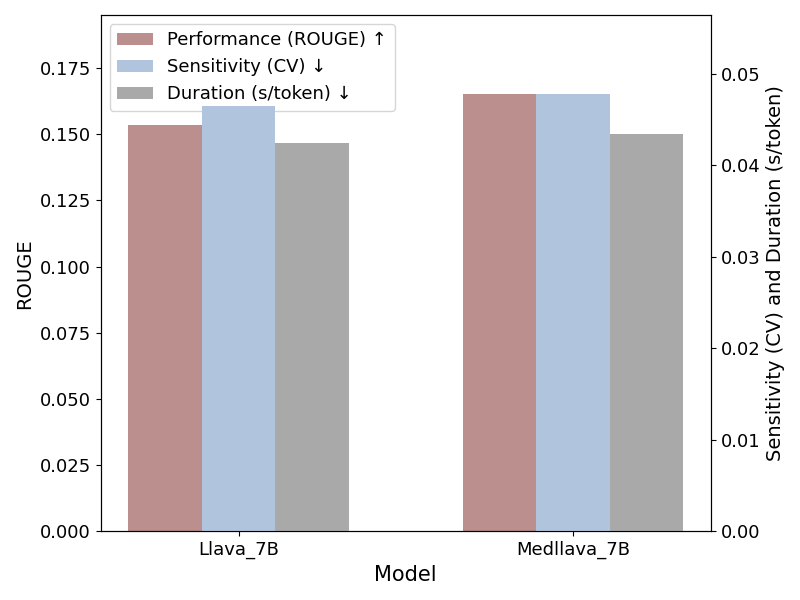}
    \caption{Image Captioning}
    \label{fig:task5}
  \end{subfigure}
  \caption{Per–model performance, measured as mean ROUGE across decoding strategies, and sensitivity, measured as the coefficient of variation (CV), for the five tasks. Model names are followed by their number of parameters and are ordered from smallest to largest; the final model is the medical LLM.}
  \label{fig:figure2}
\end{figure*}

We evaluated the models from three perspectives: performance, measured by the mean ROUGE score; sensitivity, quantified using the coefficient of variation ($CV=std/mean$) to assess consistency across decoding strategies; and inference time, measured as the number of seconds per generated token. Figure \ref{fig:figure2} presents bar charts showing these for all models divided by tasks. 

Among general models (excluding multimodal ones), we found a strong and statistically significant positive correlation between model size and task-normalized performance, with larger models achieving higher scores (Kendall’s $\tau = 0.7807$, $p = 0.0151$). A similar trend was found between model size and inference time: larger models decoded more slowly (Kendall’s $\tau = 0.6141$, $p = 0.0081$). In contrast, model size did not show a significant correlation with sensitivity. 

While general models occasionally outperformed medical domain-specific LLMs on individual tasks, this advantage was inconsistent, and no statistically significant difference in average performance was found between the two groups. Inference speed likewise showed no meaningful difference between the two groups. Sensitivity to decoding, however, did differ. We applied \textit{Levene’s} test, which assesses whether two or more groups have equal variances. Because our distributions are not normal, we also report the Brown–Forsythe modification, which substitutes the group median for the mean when computing deviations, making the test more robust. The results were consistent in both tests (Levene: $F_{(1,16)} = 20.10$, $p = 0.0004$; Brown–Forsythe: $F_{(1,16)} = 7.74$, $p = 0.013$; $n = 18$), showing that medical LLMs are more sensitive to the decoding strategy.


\subsection{Metrics}

\begin{figure*}[htb!]
  \centering
   \begin{subfigure}[t]{0.8\textwidth}
    \centering
    \includegraphics[width=\linewidth]{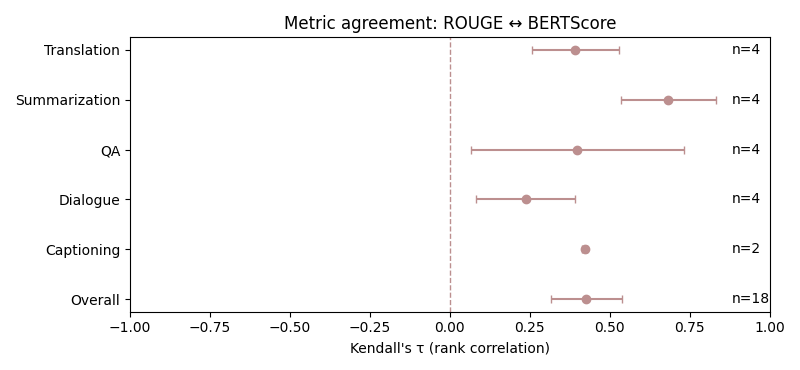}
    \vspace{-10pt}
    \caption{}
    \label{fig:figure3a}
  \end{subfigure}
  \begin{subfigure}[t]{0.8\textwidth}
    \centering
    \includegraphics[width=\linewidth]{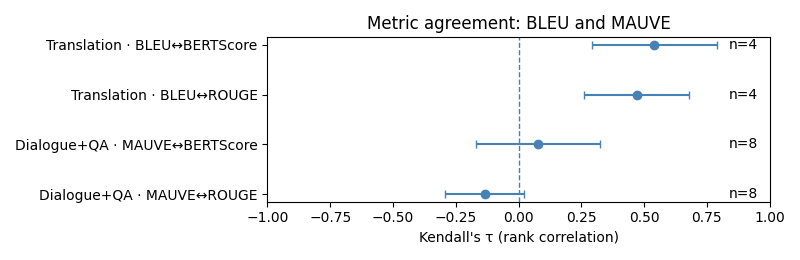}
    \vspace{-10pt}
    \caption{}

    \label{fig:figure3b}
  \end{subfigure}\hfill
 \caption{Metric agreement (Kendall’s $\tau$): (a) ROUGE–BERTScore overall and by task; (b) BLEU–BERTScore and BLEU–ROUGE (translation), and MAUVE–BERTScore and MAUVE–ROUGE (dialogue+QA).}
  \label{fig:figure3}
\end{figure*}

We measured rank agreement between metrics using Kendall’s $\tau$. Figure~\ref{fig:figure3a} shows a forest plot of ROUGE–BERTScore correlations, reported per task and overall. Error bars indicate 95\% confidence intervals across task–model blocks. The agreement is highest for summarization, followed by captioning, QA, and translation, and lowest for dialogue. Figure~\ref{fig:figure3b} shows correlations for BLEU with ROUGE and BERTScore (translation), and MAUVE with ROUGE and BERTScore (Dialogue+QA). BLEU aligns strongly with both metrics, whereas MAUVE shows weak, sometimes negative correlation with them.

Furthermore, we quantified the metric sensitivity to decoding using the CV, computed within each task$\times$model block across strategies. BERTScore was the most stable (mean CV $\approx$ 0.002), followed by BLEU ($\approx$ 0.019) and ROUGE ($\approx$ 0.029), whereas MAUVE was by far the most variable ($\approx$ 0.162). 

\section{Discussion}
\vspace{-5pt}
Deterministic decoding strategies, especially search-based ones like BS, CS, and DBS, achieved higher ROUGE scores than stochastic methods such as top-p, top-k, and eta sampling, at the cost of increased inference time. The positive correlation between decoding latency and performance highlights a trade-off in medical applications, where precision and reliability are essential. Our results indicate that the choice of decoding strategy should be regarded as an important design decision in the deployment of LLMs for medical use, as it can affect both patient safety and the quality of clinical decision support; for instance, by producing unsafe dosage recommendations or, conversely, overly generic guidance.

Interestingly, in summarization, the min-p strategy achieved the highest performance, suggesting that its adaptive balance between coherence and diversity is particularly effective in the context of medical summarization. By dynamically adjusting the probability threshold, min-p can preserve key factual content while allowing sufficient linguistic variety to produce fluent summaries. This property can be advantageous in healthcare contexts, where summaries must be both accurate and readable for diverse audiences, including clinicians and patients.

Beyond decoding, our model analysis shows that bigger models are more performant but slower, as expected; however, they are not more robust with respect to decoding sensitivity. Medical LLMs outperformed general models in only two of the five tasks, indicating that domain-specific pre-training does not guarantee performance gains. They were also slower, although size-matched comparisons are too limited for firm conclusions. Another interesting observation is that medical models are far more sensitive than general ones to the choice of decoding strategy. This means that a medical model that performs well with greedy decoding can perform much worse if the strategy is changed, whereas general models shift only marginally. Medical models are fine-tuned on a smaller, domain-specific dataset, and fine-tuning has shown to make language models less calibrated \cite{kong-etal-2020-calibrated}. Moreover, when a model is specialized in a certain domain, it tends to assign high probabilities to tokens relevant to in-domain questions, but for out-of-domain questions, the probability distribution may be flatter, which increases the influence of decoding methods and can lead to larger variations in output quality.

For statistical analysis, we used ROUGE as the primary metric to ensure a consistent comparison between decoding methods. We also included BLEU, a traditional n-gram–based measure for translation; BERTScore, which uses contextual embeddings to capture semantic similarity; and MAUVE, which measures distributional similarity and is often applied to dialogue.

Metric agreement patterns varied: BERTScore and ROUGE aligned most closely for summarization and least for dialogue, the latter expected given its open-ended nature. Translation might be expected to show high agreement due to its rigid structure, but ROUGE’s reliance on exact token matches may penalize valid paraphrases that BERTScore can capture. As anticipated, BLEU and ROUGE correlated strongly, and both correlated moderately with BERTScore. MAUVE, in contrast, showed weak or negative correlations with the others, likely because it emphasizes diversity and naturalness over reference overlap. It was also very sensitive to decoding strategy, while BERTScore was the most stable.

In the medical domain, where accuracy is essential, MAUVE alone may be insufficient; combining it with a precision-oriented metric could yield a more balanced evaluation, a direction future work should explore. A limitation of our study is the lack of analysis on repetition and frequency penalties, despite their growing use in practice. While automatic metrics are valuable for scalability and rapid experimentation, they cannot fully capture aspects such as factual correctness or user satisfaction. Thus, human evaluation continues to play a vital role alongside metric-based assessment.
\vspace{-5pt}
 
\section{Conclusion}
\vspace{-5pt}
Our findings show in summary that while deterministic, search-based decoding strategies like beam search generally produce higher quality results at the cost of increased inference time, the optimal choice is task-dependent. Notably, the impact of the decoding strategy can be as significant as the choice of the LLM itself. We also found that larger models, while more performant, are not more robust to decoding sensitivity. A surprising result is that medical-specific LLMs do not show a consistent performance advantage over general models and are significantly more sensitive to the selected decoding method. This highlights a need for careful, task-specific tuning and evaluation, especially as metrics like MAUVE show weak agreement with traditional measures and greater sensitivity to decoding choices. Future work should examine whether these findings generalize to larger datasets, as our evaluation was limited in sample size.


\begin{credits}
\subsubsection{\ackname} The research presented in this paper has benefited from the Experimental Infrastructure for Exploration of Exascale Computing (eX3), which is financially supported by the Research Council of Norway under contract 270053.

\subsubsection{\discintname}
The authors have no competing interests to declare that are
relevant to the content of this article. 

\end{credits}
%
%
%
%
\clearpage
\bibliographystyle{splncs04}  
\bibliography{ref}     






\end{document}